\documentclass[10pt,twocolumn,letterpaper]{article}

\usepackage[pagenumbers]{cvpr} 

\usepackage{graphicx}
\usepackage{amsmath}
\usepackage{amssymb}
\usepackage{booktabs}

\usepackage{xcolor}
\usepackage{nicefrac}
\usepackage{multirow}
\usepackage{algpseudocode}
\usepackage{algorithm}
\usepackage[accsupp]{axessibility}

\newcommand{\edit}[1]{#1}

\newcommand{\F}{F}

\usepackage[pagebackref,breaklinks,colorlinks]{hyperref}

\usepackage[capitalize]{cleveref}
\crefname{section}{Sec.}{Secs.}
\Crefname{section}{Section}{Sections}
\Crefname{table}{Table}{Tables}
\crefname{table}{Tab.}{Tabs.}

\def\cvprPaperID{662} 
\def\confName{CVPR}
\def\confYear{2023}

\begin{document}

\title{Adversarial Counterfactual Visual Explanations}

\author{Guillaume Jeanneret, Lo\"{i}c Simon, Fr\'ed\'eric Jurie\\
University of Caen Normandie, ENSICAEN, CNRS,  France \\
{\small \texttt{guillaume.jeanneret-sanmiguel@unicaen.fr}}}
\maketitle

\begin{abstract}
    Counterfactual explanations and adversarial attacks have a related goal: flipping output labels with minimal perturbations regardless of their characteristics. Yet, adversarial attacks cannot be used directly in a counterfactual explanation perspective, as such perturbations are perceived as noise and not as actionable and understandable image modifications. Building on the robust learning literature, this paper proposes an elegant method to turn adversarial attacks into semantically meaningful perturbations, without modifying the classifiers to explain. The proposed approach hypothesizes that Denoising Diffusion Probabilistic Models are excellent regularizers for avoiding high-frequency and out-of-distribution perturbations when generating adversarial attacks. The paper's key idea is to build attacks through a diffusion model to polish them. This allows studying the target model regardless of its robustification level. Extensive experimentation shows the advantages of our counterfactual explanation approach over current State-of-the-Art in multiple testbeds.
\end{abstract}
\section{Introduction}

The research branch of explainable artificial intelligence has yielded remarkable results, gradually opening the machine learning black boxes. The production of counterfactual explanations (CE) has become one of the promising pipelines for explainability, especially in computer vision~\cite{Rodriguez_2021_ICCV,Jeanneret_2022_ACCV,steex,Singla2020Explanation}. As a matter of fact, CE are an intuitive way to expose how an input instance can be minimally modified to steer the desired change in the model's output. More precisely, CE answers the following: \textit{what does $X$ have to change to alter the prediction from $Y$ to $Y'$?} From a user perspective, these explanations are easy to understand since they are concise and illustrated by examples. Henceforth, companies have adopted CE as an interpretation methodology to legally justify the decision-making of machine learning models~\cite{wachter2018CounterfactualExplanationsOpening}. To better appreciate the potential of CE, one may consider the following scenario: a client goes to a photo booth to take some ID photos, and the system claims the photos are invalid for such usage. Instead of performing random attempts to abide by the administration criteria, an approach based on CE could provide visual indications of what the client should fix.

The main objective of CE is to add minimalistic semantic changes in the image to flip the original model's prediction. Yet, these generated explanations must accomplish several objectives~\cite{wachter2018CounterfactualExplanationsOpening,Jeanneret_2022_ACCV,Rodriguez_2021_ICCV}. A CE must be \emph{valid}, meaning that the CE has to change the prediction of the model. Secondly, the modifications have to be \emph{sparse and proximal} to the input data, targeting to provide simple and concise explanations. In addition, the CE method should be able to generate \emph{diverse} explanations. If a trait is the most important for a certain class among other features, diverse explanations should change this attribute most frequently. Finally, the semantic changes must be \emph{realistic}. When the CE method inserts out-of-distribution artifacts in the input image, it is difficult to interpret whether the flipping decision was because of the inserted object or because of the shifting of the distribution, making the explanation unclear.

Adversarial attacks share a common goal with CE: flipping the classifier's prediction. For traditional and non-robust visual classifiers, generating these attacks on input instances creates imperceptible noise. Even though it has been shown that it contains meaningful changes~\cite{ilyas2019adversarial} and that adversarial noise and counterfactual perturbations are related~\cite{pmlr-v97-etmann19a,NEURIPS2019_7392ea4c}, adversarial attacks have lesser value. Indeed, the modifications present in the adversaries are unnoticeable by the user and leave him with no real feedback.

Contrary to the previous observations, many papers (\eg, \cite{perez2021enhancing}) evidenced that adversarial attacks toward {\em robust} classifiers generate semantic changes in the input images. This has led works~\cite{santurkar2019image,zhu2021towards} to explore robust models to produce data using adversarial attacks. In the context of counterfactual explanations, this is advantageous~\cite{boreiko2022sparse,pmlr-v130-schut21a} because the optimization will produce semantic changes to induce the flipping of the label.

Then two challenges arise when employing adversarial attacks for counterfactual explanations.
On the one hand, when studying a classifier, we must be able to explain its behavior regardless of its characteristics. 
So, a naive application of adversarial attacks is impractical for non-robust models. 
On the other hand, according to~\cite{TsiprasSETM19}, robustifying the classifier yields an implicit trade-off by lowering the \emph{clean accuracy}, as referred by the adversarial robustness community~\cite{croce2020reliable}, a particularly crucial trait for high-stakes areas such as the medical field~\cite{mertes2022ganterfactual}.

The previous remarks motivate our endeavor to mix the best of both worlds. Hence, in this paper, we propose robustifying brittle classifiers \emph{without} modifying their weights to generate CE. This robustification, obtained through a filtering preprocessing leveraging diffusion models~\cite{NEURIPS2020_4c5bcfec}, allows us to keep the performance of the classifier untouched and unlocks the production of CE through adversarial attacks.

We summarize the novelty of our paper as follows:
(i) We propose Adversarial Counterfactual Explanations, ACE in short, a novel methodology based on adversarial attacks to generate semantically coherent counterfactual explanations.
(ii) ACE performs competitively with respect to the other methods, beating previous state-of-the-art methods in multiple measurements along multiple datasets. 
(iii) Finally, we point out some defects of current evaluation metrics and propose ways to remedy their shortcomings.
(iv) To show a use case of ACE, we study ACE's meaningful and plausible explanations to comprehend the mechanisms of classifiers. We experiment with ACE findings producing actionable modifications in real-world scenarios to flip the classifier decision.

\edit{Our code and models are available on \href{https://github.com/guillaumejs2403/ACE}{GitHub}.}
\section{Related Work}

\textbf{Explainable AI.} The main dividing line between the different branches of explainable artificial intelligence stands between \textit{Ad-Hoc} and \textit{Post-Hoc} methods. The former promotes architectures that are interpretable by design~\cite{rymarczyk2021interpretable,Bohle_2022_CVPR,Bohle_2021_CVPR,Huang_2020_CVPR} while the latter considers analyzing existing models as they are. Since our setup lies among the Post-Hoc explainability methods, we spotlight that this branch splits into global and local explanations. The former explains the general behavior of the classifier, as opposed to a single instance for the latter. This work belongs to the latter. There are multiple local explanations methods, from which we highlight saliency maps~\cite{Jalwana_2021_CVPR,Wang_2020_CVPR_Workshops,Lee_2021_CVPR,8354201,Kim2022HIVE,zheng2022shap}, concept attribution~\cite{pmlr-v80-kim18d,NEURIPS2019_77d2afcb,kolek2022cartoon} and model distillation~\cite{tan2018learning,Ge_2021_CVPR}. Concisely, these explanations try to shed light on \emph{how} a model took a specific decision. In contrast, we focus on the on-growing branch of counterfactual explanations, which tackles the question: \emph{what}  does the model uses for a forecast? We point out that some novel methods~\cite{vandenhende2022making,pmlr-v97-goyal19a,wang2020scout,Wang_2021_CVPR} call themselves counterfactual approaches. Yet, these systems highlight regions between a pair of images without producing any modification.

\textbf{Counterfactual Explanations.} CE have taken momentum in recent years to explain model decisions. 
Some methods rely on prototypes~\cite{looveren2021interpretable} or deep inversion~\cite{thiagarajan2021designing}, while other works explore the benefits of other classification models for CE, such as Invertible CNNs~\cite{hvilshoj2021ecinn} and Robust Networks~\cite{boreiko2022sparse,pmlr-v130-schut21a}. A common practice is using generative tools as they give multiple benefits when producing CE. In fact, using generation techniques is helpful to generate data in the image manifold. There are two modalities to produce CE using generative approaches. Many methods use conditional generation techniques~\cite{van2021conditional,Singla2020Explanation,looveren2021interpretable} to fit what a classification model learns or how to control the perturbations. Conversely, unconditional approaches~\cite{Rodriguez_2021_ICCV,nemirovsky2020countergan,Jeanneret_2022_ACCV,shih2021GANMEXOnevsoneAttributions,zhao2018GeneratingNaturalAdversarial,Khorram_2022_CVPR} optimize the latent space vectors. 


We'd like to draw attention to Jeanneret~\etal~\cite{Jeanneret_2022_ACCV}'s counterfactual approach, which uses a modified version of the guided diffusion algorithm to steer image generation towards a desired label. This modification affects the DDPM generation algorithm itself. In contrast, while we also use DDPM, we use it primarily as a regularizer before the classifier. Instead of controlling the generation process, we generate semantic changes using adversarial attacks directly on the image space, and then post-process the image using a standard diffusion model.  Furthermore, we use a refinement stage to perform targeted edits only in regions of interest.


\textbf{Adversarial Attacks and their relationship with CE.} Adversarial attacks share the same main objective as counterfactual explanations: flipping the forecast of a target architecture. On the one hand, \textit{white-box} attacks~\cite{DBLP:journals/corr/GoodfellowSS14,madry2018towards,carlini2017towards,moosavi2016deepfool,croce2020reliable,Jeanneret_2021_ICCV} leverage the gradients of the input image with respect to a loss function to construct the adversary. In addition, universal noises~\cite{moosavi2017universal} are adversarial perturbations created for fooling many different instances. On the other hand, \textit{black-box} attacks~\cite{zhou2018transferable,poursaeed2018generative,ACFH2020square} restrain their attack by checking merely the output of the model. Finally, Nie~\etal~\cite{nie2022DiffPure} study DDPMs from a robustness perspective, disregarding the benefits of counterfactual explanations.

In the context of CE for visual models, the produced noises are indistinguishable for humans when the network does not have any defense mechanism, making them useless. This lead works~\cite{NEURIPS2019_7392ea4c,akhtar2021attack,pawelczyk2022exploring} to approach the relationship between these two research fields. Compared to previous approaches, we manage to leverage adversarial attacks to create semantic changes in undefended models to explore their semantic weaknesses perceptually in the images; a difficult task due to the nature of the data.

\begin{figure*}[!t]
    \centering
    \includegraphics[width=0.90\textwidth]{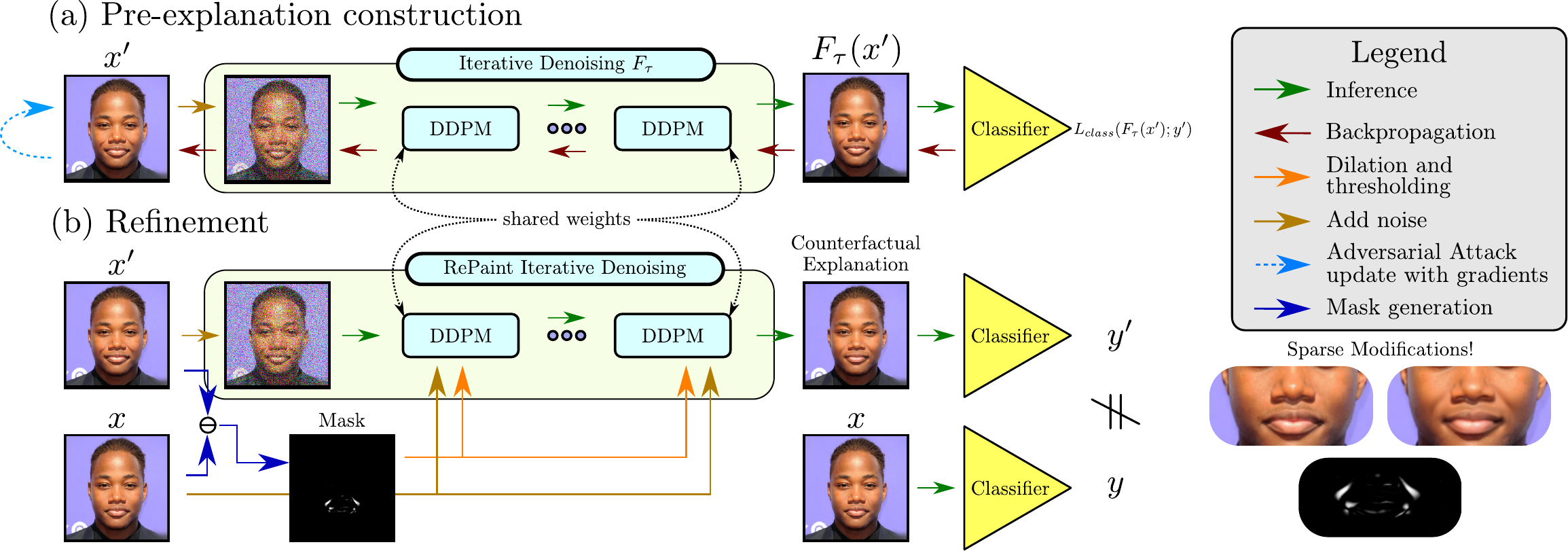}
    \caption{\textbf{Pre-explanation Construction and Refinement} ACE generates the counterfactual explanation in a two-step sequence. Initially, (a) To generate semantic updates in the input image, the DDPM processes the instance before computing the loss function $L_{class}(F_\tau(x'); y')$, where $y'$ is the target label. \edit{To simplify the process, we omit the distance loss between the perturbed image $x'$ and the input image $x$.}. 
    \edit{Then, we compute the gradients with respect to $x'$ and update it using the adversarial attack.}
    Finally, (b) we generate a binary mask using the magnitude's difference between the explanation and input image to refine the pre-explanation using RePaint's inpainting method.}
    \label{fig:ace}
\end{figure*}

\section{Adversarial Counterfactual Explanations}

The key contribution of this paper is our novel Adversarial Counterfactual Explanations (ACE) method. 
ACE produces counterfactual images in two steps, as seen in Figure~\ref{fig:ace}. We briefly introduce these two steps here and detail them in the following sections.\\
{\em Step 1. Producing pre-explanation images (\S\ref{ssec:pre-exp}).} Let $L_{class}(x; y)$ be a function measuring the agreement between the sample $x$ and class $y$. This function is typically the cross-entropy loss of the classifier we are studying with respect to $y$. With ACE, generating the pre-explanation image of $(x,y)$ for the target class $y'\neq y$ consists in finding $x'$ minimizing  $L_{class}(F(x'); y')$ \edit{using the adversarial attack as the optimizer}. Here, $F(x')$ is a filtering function that constrains the attack to stay in the manifold of the training images. In a nutshell, the filtering process $F$ robustifies the fragile classifier under examination to generate semantic changes \emph{without} modifying its weights. \\
{\em Step 2. Bringing the pre-explications closer to the input images (\S\ref{ssec:repaint}).} The pre-explanation generation restricts only those pixels in the image that are useful in switching the output label from $y$ to $y'$. The rest of the pixels are only implicitly constrained by the design of $F$. Accordingly, the purpose of this second step is to keep these non-explicitly constrained pixels identical to those of the input image.

\subsection{Pre-explanation generation with DDPMs}\label{ssec:pre-exp}

To avoid generating adversarial noise and producing useful semantics, the previously introduced function $F$ should have two key properties. (i) Removing high-frequency information that traditional adversarial attacks generate. Indeed, these perturbations could change the classifier's decision without being actionable or understandable by a human. (ii) Producing in-distribution images without distorting the input image. This property seeks to maintain the image structures not involved in the decision-making process as similar as possible while avoiding giving misleading information to the user.

Denoising Diffusion Probabilistic Models~\cite{NEURIPS2020_4c5bcfec}, commonly referred to as DDPM or diffusion models, achieve these properties if used properly. On the one hand, each inference through the DDPM is a denoising process; in particular, it removes high-frequency signals. On the other hand, DDPMs generate in-distribution images.

As a reminder, DDPMs rely on two Markov chains, one inverse to the other. The forward chain \emph{adds} noise from a state $t$ into $t+1$ while the reverse chain \emph{removes} it from $t+1$ to $t$. Noting $x_t$ the instance at time step $t$, the forward chain is directly simulated from a clean instance $x_0$ through 
\begin{equation}\label{eq:encode}
    x_t = \sqrt{\Bar{\alpha}_t} \, x_0 + \sqrt{1 - \Bar{\alpha}_t} \, \epsilon, \; \epsilon \sim \mathcal{N}(0,I),
\end{equation}
where $\Bar{\alpha}_t$ is a time-dependent constant. At inference, the DDPM produces a mean $\mu_t(x_t)$ and a deviation matrix $\Sigma_t(x_t)$. 
Using these variables, the next less noisy image is sampled from
\begin{equation}\label{eq:decode}
    x_{t-1} = \mu_t(x_t) + \Sigma_t(x_t) \, \epsilon, \; \epsilon \sim \mathcal{N}(0,I).
\end{equation}
Thus, the DDPM denoising algorithm iterates the previous step until $t=0$ arriving at an image without noise. Please refer to previous works~\cite{Dhariwal2021DiffusionMB,NEURIPS2020_4c5bcfec} for a thorough understanding of diffusion models. 

\textbf{ACE pre-explanation generation.} Starting from a query image $x$, we can obtain a filtered version by applying the forward DDPM process up to level $\tau$ (Eq.~\ref{eq:encode}) and then denoise it recursively thanks to the iterative DDPM denoising steps (Eq~\ref{eq:decode}) starting from level $t=\tau$. In this case, to highlight the use of this intermediate step $\tau$, we denote the diffusion filtering process as $F=\F_\tau$ (Figure~\ref{fig:ace}a). Thus, we optimize the image through the DDPM filtering process, $\F_\tau$, before computing the classification loss. Henceforth, we obtain the pre-explanations by optimizing 
\begin{equation}\label{eq:optim}
    \underset{x'}{\mathrm{argmin}}\; L_{class}(\F_\tau(x'); y') + \lambda_d\, d(x',x)
\end{equation}
\edit{using the adversarial attack of choice. Here,} $\lambda_d$ is a regularization constant and $d$ a distance function.

\subsection{Bringing the pre-explanations closer to the input images}
\label{ssec:repaint}

By limiting the value of $\tau$, the DDPM will not go far enough to generate a normal distribution, and the reconstruction will somehow preserve the overall structure of the image. However, we noted that a post-processing phase could help keep irrelevant parts of the image untouched. For example, in the case of faces, the denoising process may change the hairstyle while targeting the smile attribute. Since hairstyle is presumably uncorrelated with the smile feature, the post-process should neutralize those unnecessary alterations.

To this end, we first compute a binary mask $m$ delineating regions that qualify for modifications. 
To do so, we consider the magnitude difference between the pre-explanation and the original mask, we dilate this gray-scale image and threshold it, yielding the desired mask.
This matter being settled, we need to fuse the CE inside the mask along with the input outside the mask to accomplish our objective.

With that aim, a natural strategy is using inpainting methods.
So, we leverage RePaint's recent technique~\cite{Lugmayr_2022_CVPR}, originally designed for image completion, and adapt it to our {\em picture-in-picture} problem (Figure~\ref{fig:ace}b).
This adaptation straightforward and integrates very well with the rest of our framework. 
It starts from the noisy pre-explanations $x_\tau$ and iterate the following altered denoising steps:
\begin{equation}
    x_{t-1} = \mu_t(x_t') + \Sigma_t(x_t') \, \epsilon, \; \epsilon \sim \mathcal{N}(0,I),
\end{equation}
where $x'_{t} = x_t \cdot m + x^i_t \cdot (1 - m)$ is the raw collage of the current noisy reconstruction $x_t$ and the noisy version of the initial instance $x^i_t$ at the same noise level $t$, obtained with Eq.~\ref{eq:encode}. 
The final image, $x_0$, will be our counterfactual explanation -- identical to the input sample outside the mask, and very similar to the pre-explanation within the mask.
In the supplementary material, we added an overview of ACE.


\section{Experimentation}
\subsection{Evaluation Protocols and Datasets}

\begin{table*}[t]
    \centering
    \footnotesize
    \begin{tabular}{c|ccccccc|ccccccc} \toprule
        \multicolumn{1}{|c}{} & \multicolumn{7}{|c|}{\textbf{Smile}} & \multicolumn{7}{c|}{\textbf{Age}}\\\midrule
        Method        & FID  & sFID & FVA  & FS & MNAC  & CD  & COUT & FID  & sFID & FVA  & FS & MNAC  & CD  & COUT\\ \midrule
        DiVE          & 29.4 & -    & 97.3 & -  & -     & -    & -   & 33.8 & -    & \textit{98.2} & -         & 4.58 & -    & - \\ 
        DiVE$^{100}$  & 36.8 & -    & 73.4 & -  & 4.63  & 2.34 & -   & 39.9 & -    & 52.2 & -  & 4.27 & -    & - \\ 
        STEEX     & 10.2 & -    & 96.9 & -  & 4.11  & -    & -   & 11.8 & -    & 97.5 & -  & 3.44 & -    & - \\
        DiME          & 3.17 & 4.89 & 98.3 & 0.729     & 3.72  & 2.30 & 0.5259 & 4.15 & 5.89 & 95.3 & 0.6714    & \textit{3.13} & 3.27 & 0.4442 \\\midrule
        ACE $\ell_1$  & \textbf{1.27} & \textbf{3.97} & \textbf{99.9} & \textbf{0.874}     & \textit{2.94}  & \textit{1.73} & \textbf{0.7828} & \textbf{1.45} & \textbf{4.12} & \textbf{99.6} & \textit{0.7817}    & 3.20 & \textit{2.94} & \textbf{0.7176} \\
        ACE $\ell_2$  & \textit{1.90} & \textit{4.56} & \textbf{99.9} & \textit{0.867}     & \textbf{2.77}  & \textbf{1.56} & \textit{0.6235}  & \textit{2.08} & \textit{4.62} & \textbf{99.6} & \textbf{0.7971}    & \textbf{2.94} & \textbf{2.82} & \textit{0.5641} \\\bottomrule

    \end{tabular}
    \caption{\textbf{CelebA Assessment.} Main results for CelebA dataset. We extracted the results from DiME and STEEX papers. In \textbf{bold} and \textit{italic} we show the best and second-best performances. ACE outperforms all methods in every assessment protocol.}
    \label{tab:celeba-main}
\end{table*}

\noindent \textbf{Datasets.} In line with the recent literature on counterfactual images \cite{Rodriguez_2021_ICCV,Singla2020Explanation,Jeanneret_2022_ACCV,Joshi2018xGEMsGE}, first, we evaluate ACE on CelebA~\cite{liu2015faceattributes}, with images of size of $128\times128$ and a DenseNet121 classifier ~\cite{huang2017densely}, for the `smile' and `age' attributes. Following Jacob~\etal~\cite{steex}, we experimented on CelebA HQ~\cite{CelebAMask-HQ} and BDD100k~\cite{Yu2020BDD100KAD}. CelebA HQ has a higher image resolution of $256\times256$. BDD100k contains complex traffic scenes as $512\times256$ images; the targeted attribute is `forward' vs `slow down'. The decision model is also a DenseNet121, trained on the BDD-IOA~\cite{Xu2020ExplainableOA} extension dataset. Regarding the classifiers for which we want to generate counterfactuals, we took the pre-trained weights from DiME~\cite{Jeanneret_2022_ACCV} source for CelebA and from STEEX~\cite{steex} for CelebA HQ and BDD100k, for fair comparisons.

\noindent \textbf{Evaluation criteria for quantitative evaluation.}\\
{\em Validity of the explanations} is commonly measured with the Flip Rate (\underline{FR}), \ie how often the CE is classified as the targeted label. \\
{\em Diversity} is measured by extending the diversity assessment from Mothilal~\etal~\cite{Mothilal2020ExplainingML}. As suggested by Jeanneret~\etal~\cite{Jeanneret_2022_ACCV}, the diversity is measured as the average LPIPS~\cite{Zhang_2018_CVPR_Unreasonable} distance between pairs of counterfactuals (\underline{$\sigma_L$}).\\
{\em Sparsity or proximity} has been previously evaluated with several different metrics~\cite{Rodriguez_2021_ICCV,Singla2020Explanation}, in the case of face images and face attributes. On the one hand, the mean number of attributes changed (\underline{MNAC}) measures the smallest amount of traits changed between the input-explanation pair. Similarly, this metric leverages an oracle network pretrained on VGGFace2~\cite{8373813} and then fine-tuned on the dataset. Further, Jeanneret~\etal~\cite{Jeanneret_2022_ACCV} showed the limitations of the MNAC evaluation and proposed the CD metric to account for the MNAC's limitations. On the other hand, to measure whether an explanation changed the identity of the input, the assessment protocol uses face verification accuracy~\cite{8373813} (\underline{FVA}). To this end, the evaluation uses a face verification network. However, FVA has 2 main limitations: i) it can be applied to face related problems only, ii) it works at the level of classifier decisions which turns out to be too rough when comparing an image to its CE, as it involves only a minimal perturbation. For face problems, we suggest skipping the thresholding and consider the mean cosine distance between the encoding of image-counterfactual pairs, what we refer to as Face Similarity (\underline{FS}). To tackle non-face images, we propose to extend FS by relying on self-supervised learning to encode image pairs. To this end, we adopted SimSiam~\cite{Chen_2021_CVPR} as an encoding network to measure the cosine similarity. We refer to this extension as SimSiam Similarity (\underline{$S^3$}).  
Finally, also for classifiers that are not related to faces, Khorram \etal~\cite{Khorram_2022_CVPR} proposed \underline{COUT} to measure the transition probabilities between the input and the counterfactual.\\
\noindent{\em Realism of counterfactual images}~\cite{Singla2020Explanation} is usually evaluated by the research community with the \underline{FID}~\cite{NIPS2017_8a1d6947} between the original set and the valid associated counterfactuals. We believe there is a strong bias as most of the pixels of counterfactuals are untouched and will dominate the measurement, as observed in our ablation studies (Sec.~\ref{sec:ablations}). To remove this bias, we split the dataset into two sets, generating the CE for one set and measuring the FID between the generated explanations and the other set,  iterating this process ten times and taking the mean. We call this metric \underline{sFID}.

\noindent \textbf{Implementation details.} One of the main obstacles of diffusion models is transferring the gradients through all the iterations of the iterative denoising process. Fortunately, diffusion models enjoy a time-step re-spacing mechanism, allowing us to reduce the number of steps at the cost of a quality reduction. So, we drastically decreased the number of sampling steps to construct the pre-explanation. For CelebA~\cite{liu2015faceattributes}, we instantiate the DDPM~\cite{Dhariwal2021DiffusionMB} model using DiME's~\cite{Jeanneret_2022_ACCV} weights. In practice, we set $\tau=5$ out of 50 steps. For CelebA HQ~\cite{CelebAMask-HQ}, we fixed the same $\tau$, but we used the re-spaced time steps to 25 steps. For BDD100k~\cite{Yu2020BDD100KAD}, we follow the same settings as STEEX~\cite{steex}: we trained our diffusion model on the 10.000 image subset of BDD100k. To generate the explanations, we used 5 steps out of 100. Additionally, all our methods achieve a success ratio of 95\% at minimum. We will detail in the supplementary material all instructions for each model on every dataset. We adopted an $\ell_1$ or $\ell_2$ distance for the distance function. Finally, for the attack optimization, we chose the PGD~\cite{madry2018towards} without any bound and with 50 optimization steps.

\subsection{Comparison Against the State-of-the-Art}

\begin{table*}[t]
    \centering
    \footnotesize
    \begin{tabular}{c|ccccccc|ccccccc} \toprule
        \multicolumn{1}{|c}{} & \multicolumn{7}{|c|}{\textbf{Smile}} & \multicolumn{7}{c|}{\textbf{Age}}\\\midrule
        Method        & FID  & sFID & FVA  & FS & MNAC  & CD    & COUT & FID  & sFID & FVA  & FS & MNAC & CD   & COUT \\ \midrule
        DiVE          &107.0 & -    & 35.7 & -  & 7.41  & -     & -    &107.5 & -    & 32.3 & -  & 6.76 & -    & - \\ 
        STEEX     & 21.9 & -    & \textit{97.6} & -         & 5.27  & -     & - & 26.8 & -    & 96.0 & -         & 5.63 & -    & - \\
        DiME          & 18.1 & 27.7 & 96.7 & 0.6729    & 2.63  & \textbf{1.82}  & \textbf{0.6495} & 18.7 & \textit{27.8} & \textit{95.0} & 0.6597    & 2.10 & \textit{4.29} & \textbf{0.5615} \\\midrule
        ACE $\ell_1$  & \textbf{3.21} & \textbf{20.2} & \textbf{100.0}& \textbf{0.8941}    & \textbf{1.56}  & 2.61  & 0.5496 & \textbf{5.31} & \textbf{21.7} & \textbf{99.6} & \textbf{0.8085}    & \textbf{1.53} & 5.4  & 0.3984 \\
        ACE $\ell_2$  & \textit{6.93} & \textit{22.0} & \textbf{100.0}& \textit{0.8440}    & \textit{1.87}  & \textit{2.21}  & \textit{0.5946} & \textit{16.4} & 28.2 & \textbf{99.6} & \textit{0.7743}    & \textit{1.92} & \textbf{4.21} & \textit{0.5303} \\\bottomrule
    \end{tabular}
    \caption{\textbf{CelebAHQ Assessment.} Main results for CelebA HQ dataset. We extracted the results from STEEX's paper. In \textbf{bold} and \textit{italic} we show the best and second-best performances, respectively. ACE outperforms most methods in many assessment protocols.}
    \vspace{-3mm}
    \label{tab:celebahq-main}
\end{table*}
\begin{table}[t]
    \centering
    \footnotesize
    \begin{tabular}{c|ccccc} \toprule
        Method        & FID  & sFID  & S$^3$  & COUT   & FR    \\ \midrule
        \multicolumn{6}{|c|}{\textbf{BDD-OIA}} \\\midrule
        DiME          & 13.70& 26.06 & 0.9340 & 0.3188 & 91.68 \\
        ACE $\ell_1$  & \textbf{2.09} & \textbf{22.13} & \textbf{0.9980} & \textit{0.7404} & 99.91 \\
        ACE $\ell_2$  & \textit{3.3}  & \textit{22.75} & \textit{0.9949} & \textbf{0.7840} & 100.0 \\\midrule
        \multicolumn{6}{|c|}{\textbf{BDD100k}} \\\midrule
        STEEX     & 58.8 & -     & -      & -      & 99.5 \\
        DiME          & 7.94 & 11.40 & 0.9463 & 0.2435 & 90.5 \\
        ACE $\ell_1$  & \textbf{1.02} & \textbf{6.25}  & \textbf{0.9970} & \textit{0.7451} & 99.9 \\
        ACE $\ell_2$  & \textit{1.56} & \textit{6.53}  & \textit{0.9946} & \textbf{0.7875} & 99.9 \\\bottomrule
    \end{tabular}
    \caption{\textbf{BDD Assessement.} Main results for BDDOIA and BDD100k datasets. We extracted STEEX's results from their paper. In \textbf{bold} and \textit{italic} we show the best and second-best performances, respectively.}
    \vspace{-3mm}
    \label{tab:bdd-main}
\end{table}

In this section, we quantitatively compare ACE against previous State-of-the-Art methods. To this end, we show the results for CelebA~\cite{liu2015faceattributes} and CelebA HQ~\cite{CelebAMask-HQ} datasets in Table~\ref{tab:celeba-main} and Table~\ref{tab:celebahq-main}, respectively. Additionally, we experimented on the BDD100k~\cite{Yu2020BDD100KAD} dataset (Table~\ref{tab:bdd-main}). To extend the study of BDD, we further evaluated our proposed approach on the BDD-IOA~\cite{Xu2020ExplainableOA} validation set, also presented in Table~\ref{tab:bdd-main}. Since DiME~\cite{Jeanneret_2022_ACCV} showed superior performance over the literature~\cite{Rodriguez_2021_ICCV,Singla2020Explanation,Joshi2018xGEMsGE}, we compare only to DiME. 

DiME experimented originally on CelebA only. Hence, they did not tune their parameters for CelebA HQ and BBD100k. 
By running their default parameters, DiME achieves a flip rate of 41\% in CelebA HQ. We fix this by augmenting the scale hyperparameter for their loss function. DiME's new success rate is 97\% for CelebA HQ. For BDD100k, our results showed that using fewer steps improves the quality. Hence, we used 45 steps out of their re-spaced 200 steps. Unfortunately, we only managed to increase their success ratio to 90.5\%. 

These experiments show that the proposed methodology beats the previous literature on most metrics for all datasets. 
For instance, ACE, whatever the chosen distance, outmatches DiME on all metrics in CelebA. 
For the CelebA HQ, we noticed that DiME outperforms ACE only for the COUT and CD metrics. 
Yet, our proposed method remains comparable to theirs. 
For BDD100k, we remark that our method consistently outperforms DiME and STEEX. 

Two additional phenomena stand out within these results. 
On the one hand, we observed that the benefit of favoring $\ell_1$ over $\ell_2$ depends on the characteristics of the target attribute.
We noticed that the former generates sparser modifications, while the latter tends to generate broader editing. 
This makes us emphasize that different attributes require distinct modifications. 
On the other hand, these results validate the extensions for the FVA and FID metrics. 
Indeed, the difference between the FVA values on CelebA are small (from 98.3 to 99.9).  
Yet, the FS shows a major increase. 
Further, for the Age attribute on CelebA HQ, ACE $\ell_2$ shows a better performance than DiME for the FID metric. 
The situation is reversed with sFID as DiME is slightly superior.

To complement our extensive experimentation, we tested ACE on a small subset of classes on ImageNet~\cite{deng2009imagenet} with a ResNet50. 
We selected three pairs of categories for the assessment, and the task is to generate the CE targeting the contrary class. 
For the FID computation, we used only the instances from both categories but not external data since we are evaluating the in-class distribution.

We show the results in Table~\ref{tab:imagenet-main}. Unlike the previous benchmarks, ImageNet is extremely complex and the classifier needs multiple factors for the decision-making process. Our results reflect this aspect. 
We believe that current advancements in CE still need an appropriate testbed to validate the methods in complex datasets such as ImageNet. For instance, the model uses the image's context for forecasting. So, choosing the target class without any previous information is unsound. 

\subsection{Diversity Assessment}

In this section, we explore ACE's ability to generate diverse explanations. 
Diffusion models are, by design,  capable of generating distributions of images.
Like \cite{Jeanneret_2022_ACCV}, we take advantage of the stochastic mechanism to generate perceptually different explanations by merely changing the noise for each CE version.
Additionally, for a fair comparison, we do not use the RePaint's strategy here because DiME does not have any local constraints and can, as well, change useless structures, like the background. 
To validate our approach, we follow \cite{Jeanneret_2022_ACCV} assessment protocol.
Numerically, we obtain a diversity score of $\sigma_L=0.110$ while DiME reports 0.213. 
Since DiME corrupts the image much more than ACE, the diffusion model has more opportunities to generate distinct instances. 
In contrast, we do not go deep into the forward noising chain to avoid changing the original class when performing the filtering. 

To circumvent the relative lack of diversity, we vary the re-spacing at the refinement stage and the sampled noise. 
Note that later in the text, we show that using all steps without any re-spacing harms the success ratio. 
So, we set the new re-spacing such that it respects the accuracy of counterfactuals and fixed the variable number of noise to maintain the ratio between $\tau$ and the re-spaced number of sampling steps ($\nicefrac{5}{50}$ in this case). 
Our  diversity score is then of 0.1436. 
Nevertheless, DiME is better than ACE in terms of diversity, but this is at the expense of the other criteria, because its diversity comes, in part, from regions of the images that should not be modified (for example, the background).

\subsection{Qualitative Results}

\begin{table}[t]
    \centering
    \footnotesize
    \begin{tabular}{c|ccccc} \toprule
        Method        & FID  & sFID  & S$^3$  & COUT    & FR \\ \midrule
        \multicolumn{6}{|c|}{\textbf{Zebra -- Sorrel}} \\\midrule
        ACE $\ell_1$  & 84.5 & 122.7 & 0.9151 & -0.4462 & 47.0\\
        ACE $\ell_2$  & 67.7 & 98.4  & 0.9037 & -0.2525 & 81.0\\\midrule
        \multicolumn{6}{|c|}{\textbf{Cheetah -- Cougar}} \\\midrule
        ACE $\ell_1$  & 70.2 &100.5 & 0.9085 & 0.0173   & 77.0 \\
        ACE $\ell_2$  & 74.1 &102.5 & 0.8785 & 0.1203   & 95.0 \\\midrule
        \multicolumn{6}{|c|}{\textbf{Egyptian Cat -- Persian Cat}} \\\midrule
        ACE $\ell_1$  & 93.6 & 156.7& 0.8467 & 0.2491   & 85.0 \\
        ACE $\ell_2$  & 107.3& 160.4& 0.7810 & 0.3430   & 97.0 \\\bottomrule
    \end{tabular}
    \caption{\textbf{ImageNet Assessement.} We test our model in ImageNet. We generated the explanations for three sets of classes. Producing CE for these classes remains a challenge.}
    \vspace{-3mm}
    \label{tab:imagenet-main}
\end{table}

\begin{figure*}[t]
 \centering
 \includegraphics[width=0.95\textwidth]{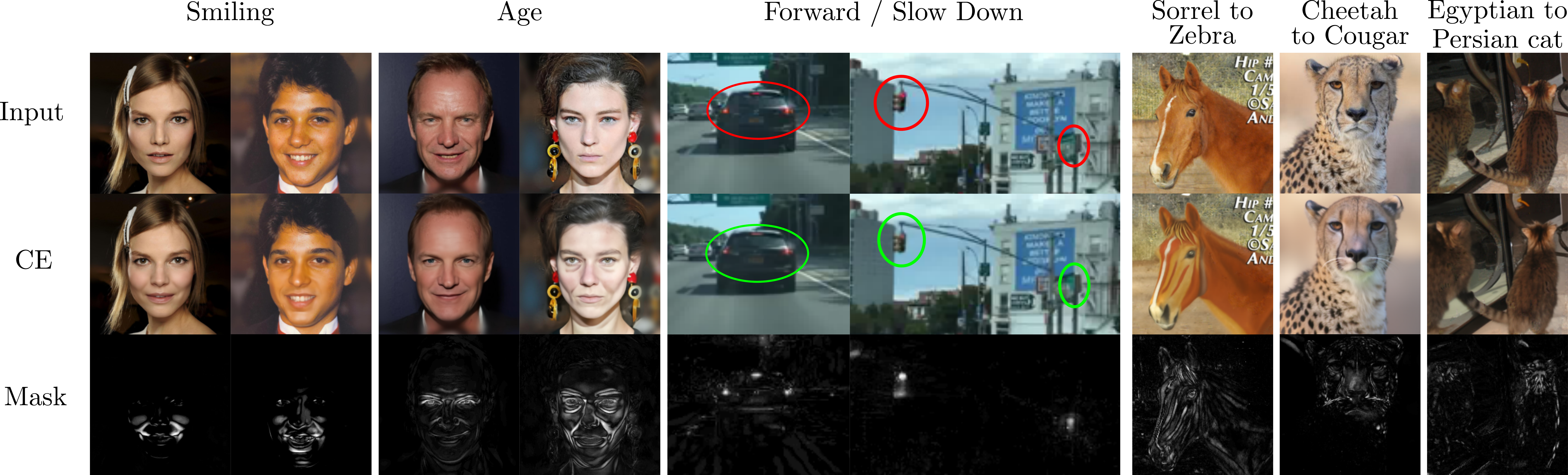}
 \caption{\textbf{Qualitative Results.} ACE create sparse but realistic changes in the input image. Further, ACE enjoys from the generate mask, which helps in understanding which and where semantic editing were added. The first row displays the input images, the second one the counterfactual explanations and the third the corresponding mask.}
 \label{fig:qualitative}
 \vspace{-3mm}
\end{figure*}

We show some qualitative results in Figure~\ref{fig:qualitative} for all datasets, included some ImageNet examples. From an attribute perspective, some have sparser or coarser characteristics. For instance, age characteristics cover a wider section of the face, while the smile attribute is mostly located in small regions of the image. Our qualitative results expose that different distance losses impose different types of explanations. For this case, $\ell_1$ loss exposes the most local and concrete explanations. On the other hand, the $\ell_2$ loss generates  coarser editing. This feature is desired for certain classes, but it is user-defined. 
Additionally, we note that the generated mask is useful to spot out the location of the changes. 
This is advantageous as it exemplifies which changes were needed and where they were added. 
Most methods do not indicate the localization of the changes, making them hard to understand. 
In the supplementary material, we included more qualitative results.

\subsection{Actionability}

\begin{figure}
 \centering
 \includegraphics[width=0.35\textwidth]{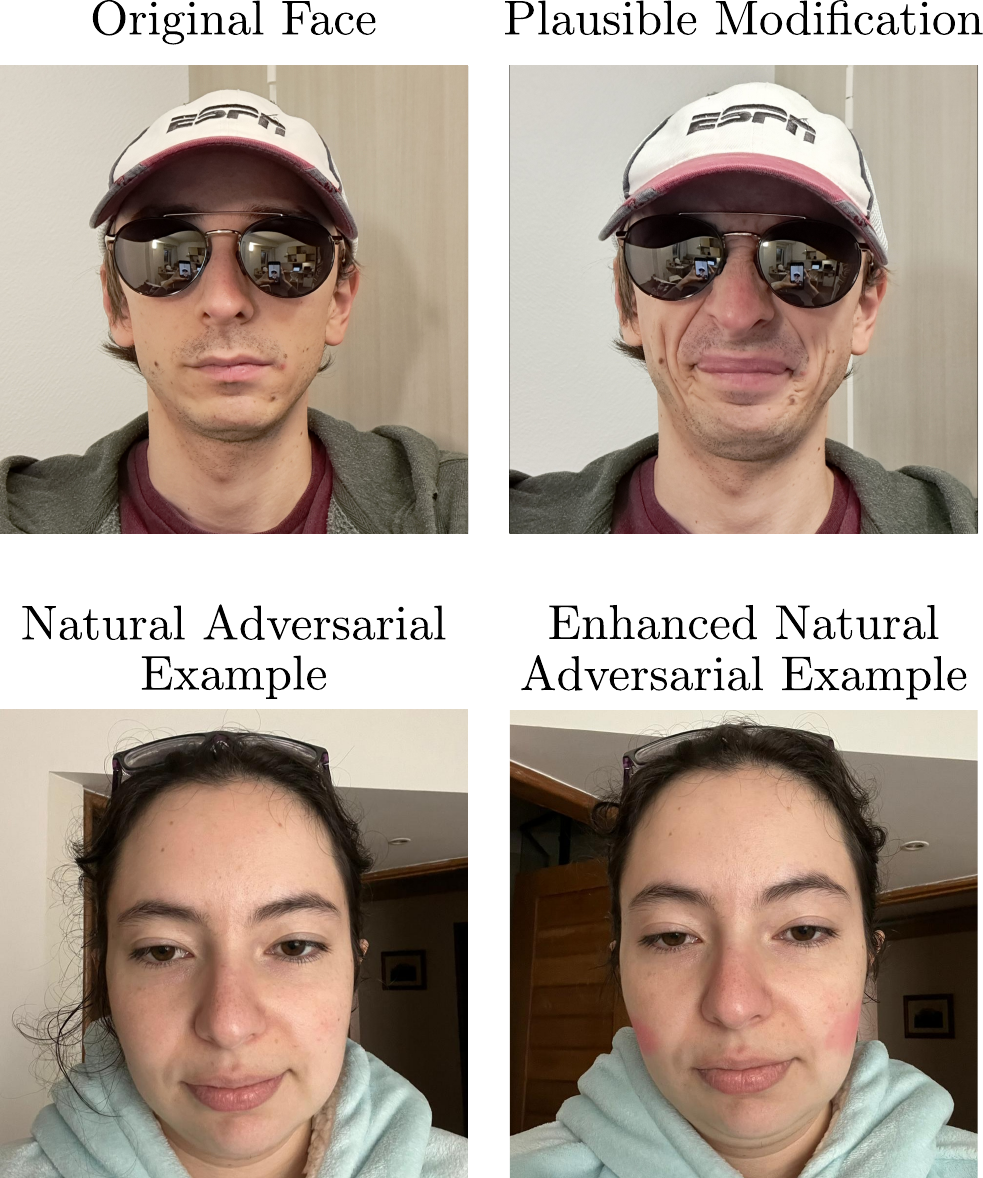}
 \caption{\textbf{Actionability.} From browsing our counterfactuals, we found two weaknesses of the scrutinized classifier. Row 1: We tested if a frown could change the classification from young to old. Row 2: we checked if having high cheekbones flipped is enough to classify someone as smiling. Both experiments were successful.}
 \label{fig:action}
 \vspace{-3mm}
\end{figure}

Counterfactual explanations are expected to teach the user plausible modifications to change the classifier's prediction. 
In this section, we study a batch of counterfactual-input tuples generated with our method. 
If ACE is capable of creating useful counterfactual explanations, we should be qualified to understand some weaknesses or some behaviors of our classifier. 
Additionally, we should be able to fool the classifier by creating the necessary changes in real life. To this end, we studied the CelebA HQ classifier for the age and smile attributes.

After surveying some images and their explanations, we identified two interesting results (Figure~\ref{fig:action}). 
Many of the counterfactual explanations changing from `young' to `old' evidence that frowning could change the prediction of the classifier. 
So, we tested this hypothesis in the real life. 
We took a photo one individual before and after the frown, avoiding changing the scenery.
We were successful and managed to change the prediction of the classifier. 
For smile, we identified a spurious correlation. 
Our counterfactuals show that the classifier uses the morphological trait of high cheekbones to classify someone as smiling as well as having red cheeks. 
So, we tested whether the classification model wrongly predicts as smiling someone with high cheekbones  even when this person  is not smiling. We also tested whether we can enhance it with some red make up in the cheeks. 
Effectively, our results show that having high cheekbones is a realistic adversarial feature toward the smiling attribute for the classifier. Also, the classifier confidence (probability) can be strengthened by adding some red make up in the cheeks. 
These examples demonstrate the applicability of ACE in real scenarios.

\subsection{Ablation Studies}\label{sec:ablations}

In this section, we scrutinize the differences between the pre-explanation and the refined explanations. Then we  explore the effects of using other types of adversarial attacks. Finally, we show that the $S^3$ metric  gives similar results as the FVA, as a sanity check.

\noindent \textbf{Pre-Explanation \textit{vs} Counterfactual Explanations.}
We explore here, quantitatively and qualitatively, the effects of the pre-explanations (Pre-CE). 
Also, we apply the diffusion model for the explanations with and without the re-spacing method, using no inpainting strategy, referred as FR-CE and F-CE, respectively.
Finally, we compare them against the complete model (ACE). 
To quantitatively compare all versions, we conducted this ablation study on the CelebA dataset for both `smile' and `age' attributes. 
We assessed the components using the FID, sFID, MNAC, CD, and FR metrics. 
We did not include the FVA or FS metrics, as these values did not vary much and do not provide insightful information; the FVA is $\sim$99.9 and FS $\sim$0.87 for all versions. 

We show the results in Table~\ref{tab:ablation-filtering}. 
We observe that pre-explanations have a low FID. 
Nonetheless, their sFID is worse than the F-CE version. 
As said before, we noticed that including both input and counterfactual in the FID assessment introduces a bias in the final measurement, and this experiment confirms this phenomenon. 
Additionally, one can check that the MNAC metric between the pre-explanation and the FR-CE version does not vary much, yet, the CD metric for the FR-CE is much better. 
This evidences that the generative model can capture the dependencies between the attributes. 
Also, we notice that the flip rate (FR) is much lower when using all diffusion steps instead of the re-spaced alternative. 
We expected this behavior, since we create the pre-explanation to change the classifier's prediction with re-spaced time steps within the DDPM.

\begin{figure}[t]
 \centering
 \includegraphics[width=0.45\textwidth]{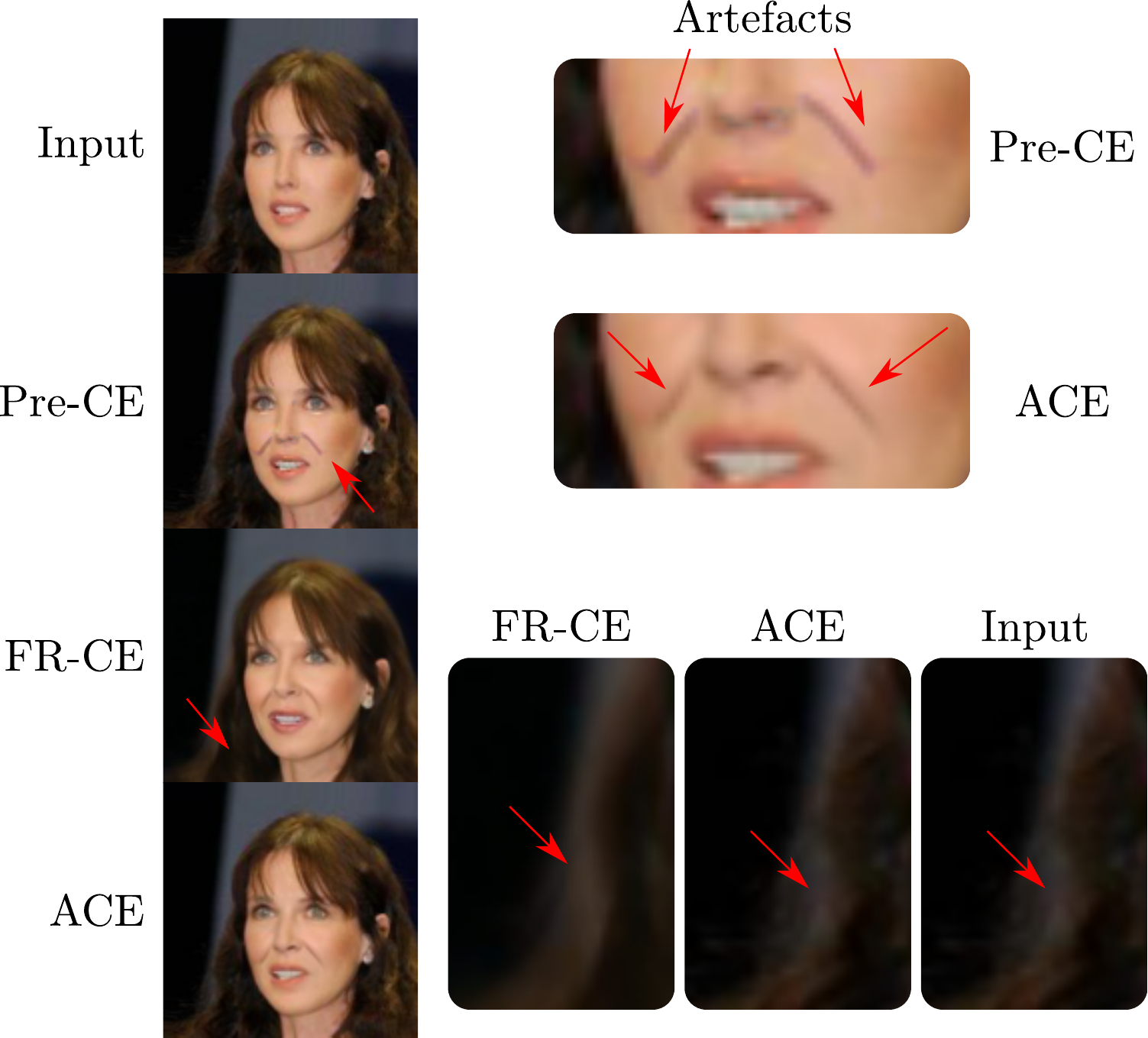}
 \caption{\textbf{Refinement Ablation.} We observe that  pre-explanations can have out-of-distribution artifacts. After filtering them, the diffusion process creates in-distribution data, but there are unnecessary changes such as the background. ACE is capable of changing the key features while avoiding modifying unwanted structures.}
 \label{fig:abl-filt}
\end{figure}

Qualitatively, we point out to Figure~\ref{fig:abl-filt}, where we exemplify the various stages of ACE. 
For instance, we see that the pre-explanation contains out of distribution artifacts and how the refinement sends it back to the image distribution. 
Also, we highlight that the filtering modifies the hair, which is not an important trait for the classifier. 
The refinement is key to avoid editing these regions.

\begin{table}[t]
    \centering
    \footnotesize
    \begin{tabular}{c|ccccc} \toprule
        \multicolumn{6}{|c|}{\textbf{Smile}} \\\midrule
        Method        & FID  & sFID & MNAC  & CD   & FR    \\ \midrule
        Pre-CE        & 1.87 & 4.63 & 3.48  & 3.05 & 99.82 \\
        FR-CE         & 8.31 & 10.30& 3.43  & 1.68 & 99.97 \\
        F-CE          & 2.64 & 4.61 & 3.16  & 1.56 & 93.37 \\ \midrule
        ACE          & 1.27 & 3.97 & 2.94  & 1.73 & 99.86 \\ \midrule
        \multicolumn{6}{|c|}{\textbf{Age}} \\\midrule
        Pre-CE        & 3.93 & 6.71 & 3.76  & 3.17 & 99.55 \\
        FR-CE         & 7.10 & 9.09 & 3.13  & 2.66 & 99.77 \\
        F-CE          & 4.23 & 6.20 & 3.53  & 3.04 & 93.50 \\ \midrule
        ACE          & 2.08 & 4.62 & 2.94  & 2.82 & 99.35 \\ \bottomrule
    \end{tabular}
    \caption{\textbf{Refinement Ablation.} We show the importance of each component from ACE. FR stands for flip rate.}
    \vspace{-3mm}
    \label{tab:ablation-filtering}
\end{table}

\begin{table}[t]
    \centering
    \footnotesize
    \begin{tabular}{c|cccc} \toprule
        Metric  & Random     & ACE        & Pre-CE     & DiME  \\ \midrule
        \multicolumn{5}{|c|}{\textbf{Smile}}\\\midrule
        FS          & 0.2649 (4) & 0.8941 (2) & 0.9200 (1) & 0.6729 (3) \\
        $S^3$       & 0.4337 (4) & 0.9876 (2) & 0.9927 (1) & 0.9396 (3) \\\midrule
        \multicolumn{5}{|c|}{\textbf{Age}}\\\midrule
        FS          & 0.2649 (4) & 0.7743 (2) & 0.8300 (1) & 0.6597 (3) \\
        $S^3$       & 0.4337 (4) & 0.9417 (2) & 0.9870 (1) & 0.9379 (3) \\\bottomrule
    \end{tabular}
    \caption{\textbf{$S^3$ equivalence to FS.} The $S^3$ metric and the FS are equivalent in a similar context. We show the metric and the order (in parentheses) and observe that both orderings are equal.}
    \label{tab:s3}
    \vspace{-3mm}
\end{table}

\noindent\textbf{Effect of Different Adversarial Attacks.}
At the core of our optimization, we have the PGD attack. 
PGD is one of the most common attacks due to its strength. 
In this section, we explore the effect of incorporating other attacks.
Thus, we tested C\&W~\cite{carlini2017towards} and the standard gradient descent (GD). 
Note that the difference between PGD and GD is that GD does not apply the $sign$ operation. 

Our results show that these attacks are capable of generating semantic changes in the image. 
Although these are as successful as the PGD attack, we require optimizing the pre-explanation for twice as many iterations. 
Even when our model is faster than~\cite{Jeanneret_2022_ACCV} \edit{--3.6 times faster--}, we require about 500 DDPM iterations to generate an explanation.

\noindent\textbf{Validity of the $S^3$ Metric.}
In this experiment, we show that the $S^3$ and the FS metrics are equivalent when used in the same test bed, \ie, CelebA HQ. 
To this end, we assess whether the ordering between ACE, pre-explanation, and DiME are equal. 
To have a reference value, we evaluate the measurements when using a pair of random images.
So, we show the values (ordering) for both metrics in Table~\ref{tab:s3} for the Age and Smiling attribute. 
As we expect, the ordering is similar between both metrics. 
Nevertheless, we stress that FS is adequate for faces since the network was trained for this task.

\section{Conclusion}

In this paper, we proposed ACE, an approach to generate counterfactual explanations using adversarial attacks. ACE relies on DDPMs to enhance the robustness of the target classifier, allowing it to create semantic changes through adversarial attacks, even when the classifier itself is not robust. ACE has multiple advantages regarding previous literature, notably seen in the counterfactual metrics. Moreover, we highlight that our explanations are capable of showing natural feature to find sparse and actionable modifications in real life, a characteristic not presented before. For instance, we were able to fool the classifier with real world changes, as well as finding natural adversarial examples. 

\edit{\textbf{Acknowledgements} 
Research reported in this publication was supported by the Agence Nationale pour la Recherche (ANR) under award number ANR-19-CHIA-0017.}


{\small
\bibliographystyle{ieee_fullname}
\bibliography{egbib}
}

\vspace{1cm}
\appendix

{\centering \textbf{\LARGE Supplementary Material
}}

\section{Detailed Implementation Details}

For each dataset, we used different configurations in architecture and for the generation of the pre-explanation. 
\edit{Yet, we tune all hyperparameters from a empirical perspective\footnote{Note that all these hyperparameters are not the same as the classically found in machine learning. These variables can be adjusted by the user in an `online' manner according to his/her expectations. Hence, a global configuration is a mere rough estimate of these parameters and can be accommodated instance-wise.}. 
We tuned $\tau$ such that the input image and its filtered instance are visually similar. Additionally, the classification between these two images are the same.
To adjust the hyperparameter $\lambda_d$, we performed a simple visual inspection.
Finally, for the threshold, we ablated its values empirically for each dataset.}
When using the distance loss $\ell_1$, we set the distance regularization constant to $\lambda_d = 0.001$ while $\lambda_d = 0.1$ for $\ell_2$.
For the final refinement, firstly, we normalize the mask by the maximum pixel's difference magnitude. 
For the dilation step, we set the mask as a square with a width and height of 15 pixels for all datasets. 
Finally, we used the cross entropy for all experiments as the $L_{class}$ loss.  
Next, we will show all implementation details for each dataset.

\textbf{CelebA}~\cite{liu2015faceattributes}: We used the same architecture and weights as~\cite{Jeanneret_2022_ACCV}. 
Additionally, we set $\tau=5$ with a total amount of steps as $50$. 
At the refinement stage, we used the same threshold of 0.15 for both $\ell_1$ and $\ell_2$ experiments for smile and age attributes.

\textbf{CelebA HQ}~\cite{CelebAMask-HQ}: Our model follows the same architecture than\cite{Dhariwal2021DiffusionMB} for ImageNet $256\times256$ unconditional generation. 
Since CelebA HQ is far less complex than ImageNet, we reduced the number of channels from 256 to 128. 
Also, our model generates samples using $500$ diffusion steps instead of $1000$. 
For training, we iterated our model for $120.000$ iterations with a batch size of 256 on two V100 GPUs following\cite{Dhariwal2021DiffusionMB}'s code. We set the learning rate to $10^4$, a weight decay of $0.05$, and no dropout.

To generate the pre-explanations, we noise the image until $\tau=5$ out of $25$ re-spaced steps. To binarize the mask, we used a threshold of $0.15$ and $0.1$ for the smiling attribute with the $\ell_1$ and $\ell_2$ distance losses, respectively. For the age attribute, we used $0.15$ for $\ell_1$ and $0.05$ for $\ell_2$.

\textbf{BDD100k/OIA}~\cite{Yu2020BDD100KAD,Xu2020ExplainableOA}: The counterfactual explanation research community opted to use BDD100k in a $512\times256$ setup. This is highly demanding computationally to create a DDPM. Thus, since we knew \textit{a priori} that we do not need many iterations for ACE to generate counterfactuals, we trained our diffusion model partially in the Markov chain. That is, our DDPM cannot generate images from pure noise. Instead, we trained it to generate images solely from a quarter of the complete chain, requiring an input instance to warm up the generation. So, we trained our model to generate instances with 250 steps out of 1000. 
This enabled us to use a lighter model. Artitecnologically, our UNet model has four downsampling stages with $128\,s$ channels, where $s$ is the downsampling stage. Finally, we used the attention layer at the deeper layer of the UNet. 
At the training phase, we used a batch size of $256$, a learning rate of $10^4$, and a weight decay and dropout of $0.05$ for $50.000$ iterations. 

To generate our explanations, we used $5$ out of $100$ (re-spaced) diffusion steps. For $\ell_1$, we used a threshold of 0.05 and 0.1 for $\ell_2$ for both datasets.

\textbf{ImageNet}~\cite{deng2009imagenet}: For this dataset, we took advantage of previous works. In this case, we utilised~\cite{Dhariwal2021DiffusionMB}'s model on ImageNet 256. To generate the explanations, we used 5 steps out of 25 for the pre-explanations and set the threshold to 0.15 to binarize the mask for all cases.

\edit{
\section{Overview of ACE}
ACE is a two-step method: firstly is the pre-explanation construction -- Algorithm~\ref{alg:pe} -- and then the refinement process -- Algorithm~\ref{alg:ce}.
To generate the pre-explanation, \textbf{(1)}~we add noise to the input image $x$ using the forward Markov chain until an intermediate step $\tau$, \ie it doesn't begin from random Gaussian noise. 
Instead, it warms up the generation with the input image through 
\begin{equation*}
    x_t = \sqrt{\Bar{\alpha}_t} \, x + \sqrt{1 - \Bar{\alpha}_t} \, \epsilon, \; \epsilon \sim \mathcal{N}(0,I).
\end{equation*}
\textbf{(2)}~ACE iteratively denoises the noisy image using the DDPM algorithm with
\begin{equation*}
    x_{t-1} = \mu_t(x_t) + \Sigma_t(x_t) \, \epsilon, \; \epsilon \sim \mathcal{N}(0,I),
\end{equation*}
where $\mu_t$ and $\Sigma_t$ are the output of the diffusion model. 
\textbf{(3)}~The scrutinized classifier uses the filtered image to compute loss function. Then, we calculate the gradients with respect to the input image $x$ in step 1, all the way through the $\tau$ steps of the diffusion model. 
\textbf{(4)}~ACE applies the gradients as the update step with the attack of choice. It iterates these four steps to create the pre-explanation.
For the refinement, it creates the mask $m$ using the difference between the pre-explanation and the original input. Then, it dilates and thresholds it to generate the binary version. 
Finally, ACE builds on RePaint to keep untouched any region lying outside the mask. The final result is the counterfactual explanation.
}

\begin{algorithm}
\caption{Pre-explanation generation}\label{alg:pe}
\begin{algorithmic}[1]
\Require Diffusion Model $D$, Distance loss $d$ and its regularization constant $\lambda_d$, classification loss $L_{class}$ comprising the classifier under observation, number of noising steps $\tau$, attack optimization algorithm $PGD$, number of update iterations $n$, initial instance $x$, target label $y$
\Function{Pre-explanation}{x, y}
\State $n \leftarrow 0$
\State $x_{orig} \leftarrow x$
\While{$n < N$}
\Comment{Attack iteration steps}
\State $\epsilon \sim \mathcal{N}(0, I)$
\State $x' \leftarrow \sqrt{\Bar{\alpha}_\tau} x + \sqrt{1 - \Bar{\alpha}_\tau} \epsilon$
\Comment{Add noise}
\State $ts \leftarrow \tau - 1$
\While{$ts \geq 0$}
\Comment{DDPM denoising}
\State $\mu, \Sigma \leftarrow D(x', ts)$
\State $\epsilon \sim \mathcal{N}(0, I)$
\State $x' \leftarrow \mu + \epsilon \Sigma$
\State $ts \leftarrow ts - 1$
\EndWhile
\State $g \leftarrow \nabla_{x'} L_{class}(x'; y') + \lambda_d d(x',x_{orig})$
\State $x \leftarrow PGD(x, g)$
\Comment{Update with attack}
\State $n + 1 \leftarrow n$ 
\EndWhile
\State \Return $x'$
\Comment{Pre-explanation}
\EndFunction
\end{algorithmic}
\end{algorithm}

\begin{algorithm}
\caption{Post-processing}\label{alg:ce}
\begin{algorithmic}[1]
\Require Diffusion Model $D$, number of noising steps $\tau$, mask dilation size $d$, threshold $u$, initial instance $x$, pre-explanation $x'$
\Function{Post-processing}{x, x'}
\State $x_{orig} \leftarrow x$
\State $\epsilon \sim \mathcal{N}(0, I)$
\State $x' \leftarrow \sqrt{\Bar{\alpha}_\tau} x' + \sqrt{1 - \Bar{\alpha}_\tau} \epsilon$
\State $ts \leftarrow \tau - 1$
\State \# Mask generation
\State $m \leftarrow sum\_over\_channels(abs(x - x'))$
\State $m \leftarrow \nicefrac{m}{maximum(m)}$
\State $m \leftarrow dilation(m, size=d) > u$
\While{$ts \geq 0$}
\Comment{DDPM denoising}
\State $\epsilon \sim \mathcal{N}(0, I)$
\State $x_{ts} \leftarrow \sqrt{\Bar{\alpha}_{ts}} x + \sqrt{1 - \Bar{\alpha}_{ts}} \epsilon$
\State $x' \leftarrow m\,x' + (1 - m)\,x_{ts}$
\State $\mu, \Sigma \leftarrow D(x', ts)$
\State $\epsilon \sim \mathcal{N}(0, I)$
\State $x' \leftarrow \mu + \epsilon \Sigma$
\State $ts \leftarrow ts - 1$
\EndWhile
\State \Return $x'$
\Comment{Counterfactual explanation}
\EndFunction
\end{algorithmic}
\end{algorithm}

\section{Qualitative Results}

\begin{figure*}[t]
    \centering
    \includegraphics[width=0.85\textwidth]{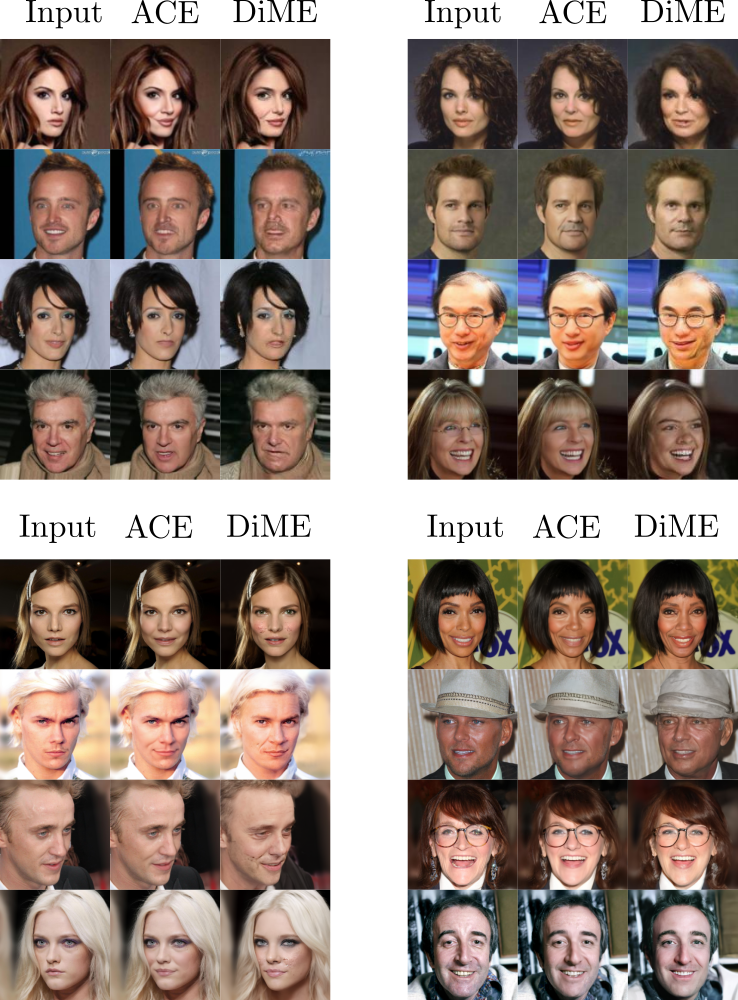}
    \caption{ACE \textit{vs.} DiME. We display some examples showing some differences between DiME counterfactuals and ACE's. In short, ACE is capable of not modifying useless information, such as the background, to generate its counterfactuals. Top row: CelebA. Bottom row: CelebA HQ. Left Column: Smiling attribute. Right Column: Age attribute.}
    \label{soup:fig:acevsdime}
\end{figure*}

In this section, we show more qualitative results. We will display the input image, its pre-explanation, the mask, and the final counterfactual for both $\ell_1$ and $\ell_2$ losses on all datasets. Note that we added a small discussion on the caption analyzing the results. In Fig.~\ref{soup:fig:acevsdime}, we compare a few examples of DiME and ACE.

\begin{figure*}[t]
    \centering
    \includegraphics[width=0.9\textwidth]{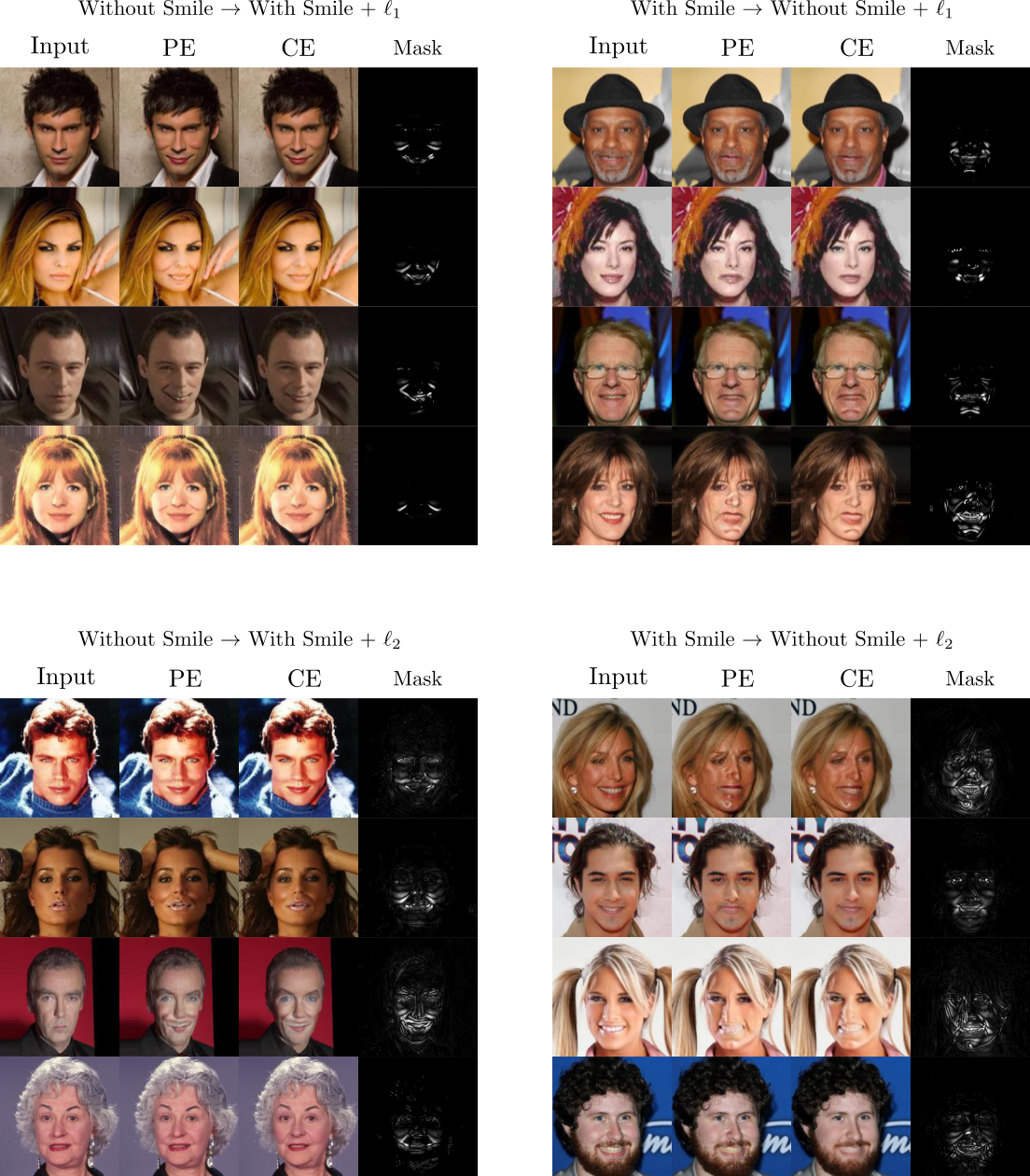}
    \caption{Additional CelebA qualitative results. We show examples for the \emph{Smiling} attribute for both distances losses. From our qualitative experiments, we see that removing the smile attributes is harder than adding them. Additionally, we see that the $\ell_1$ loss creates more sparse editings.}
    \label{soup:fig:celeba-smile}
\end{figure*}

\begin{figure*}[t]
    \centering
    \includegraphics[width=0.9\textwidth]{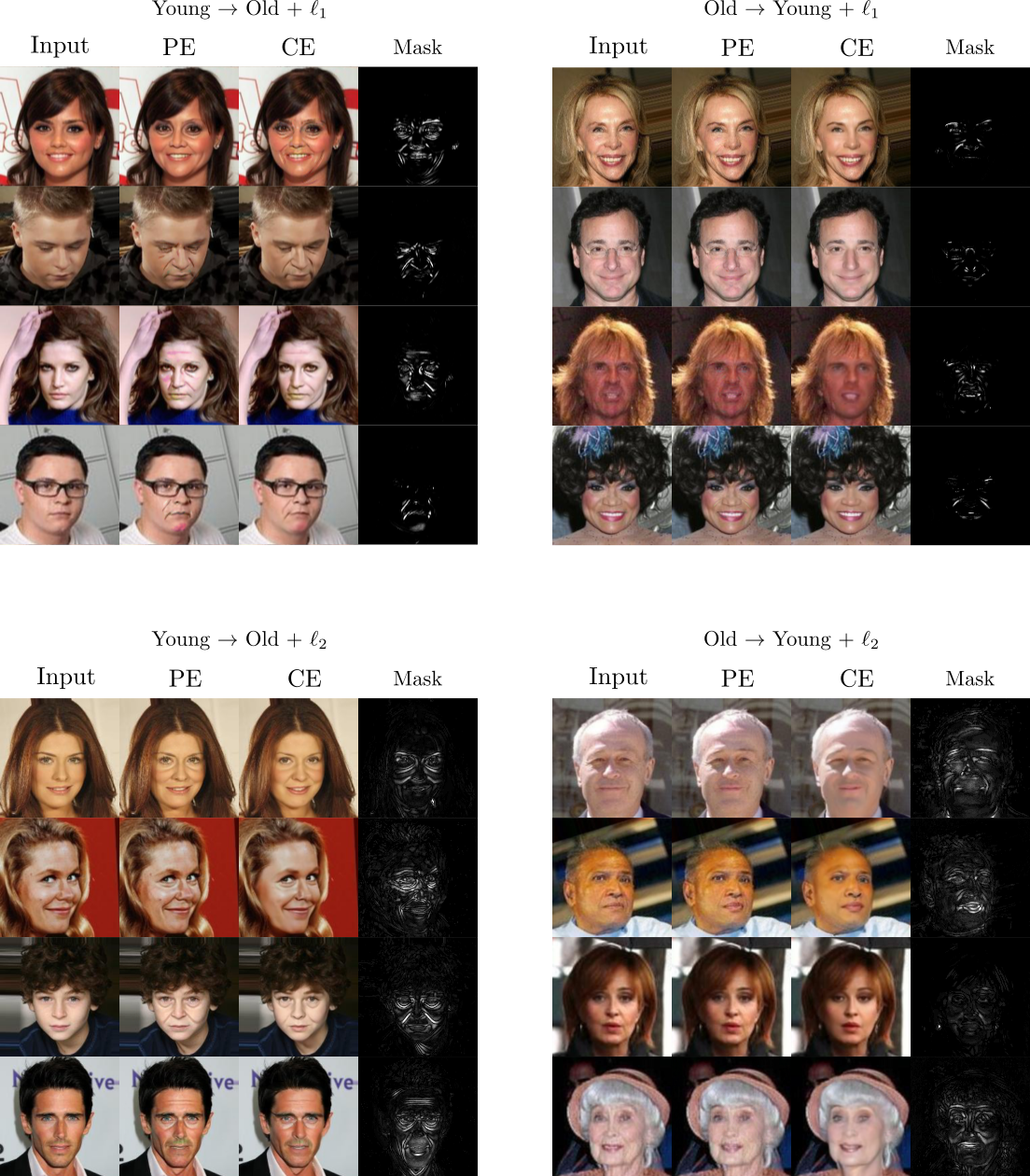}
    \caption{Additional CelebA qualitative results. We show examples for the \emph{Age} attribute for both distances losses. The results show that the $\ell_1$ loss creates more out-of-distribution artifacts.}
    \label{soup:fig:celeba-smile}
\end{figure*}

\begin{figure*}[t]
    \centering
    \includegraphics[width=0.9\textwidth]{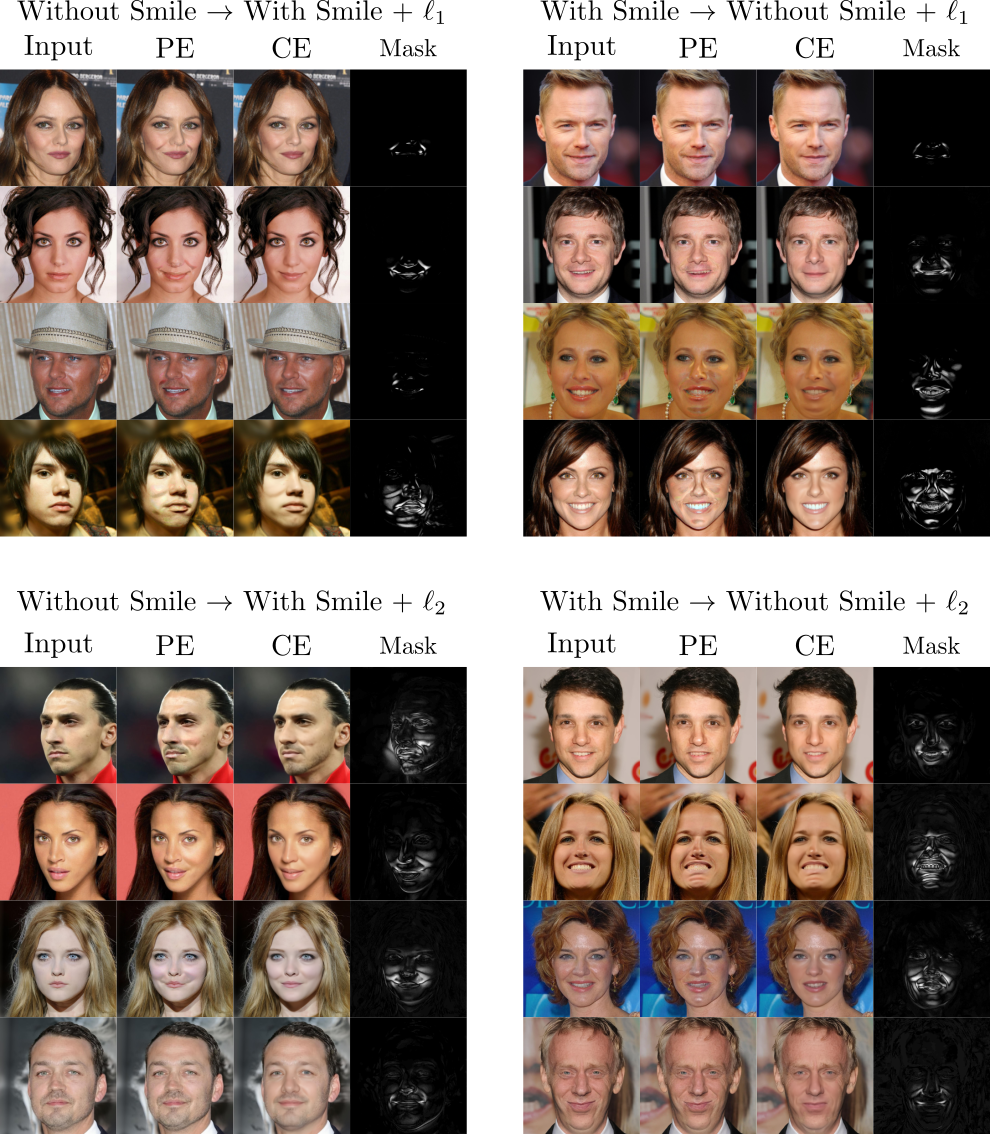}
    \caption{Additional CelebA HQ qualitative results. We show examples for the \emph{Smiling} attribute for both distances losses. We see similar behavior in the CelebA dataset.}
    \label{soup:fig:celeba-smile}
\end{figure*}

\begin{figure*}[t]
    \centering
    \includegraphics[width=0.9\textwidth]{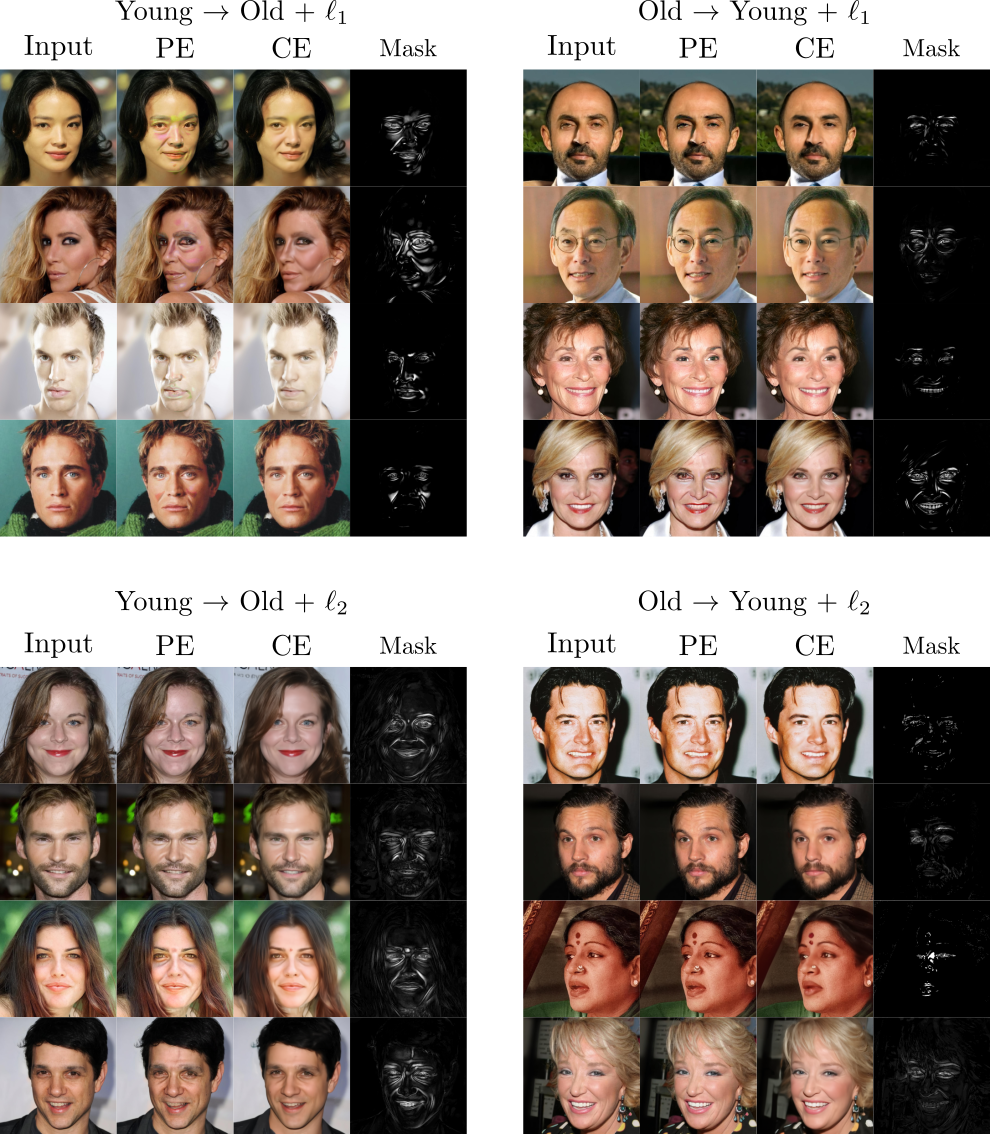}
    \caption{Additional CelebA HQ qualitative results. We show examples for the \emph{Age} attribute for both distances losses. These examples show that transforming \textit{Old} to \textit{Young} is less informative than the other way.}
    \label{soup:fig:celeba-smile}
\end{figure*}

\begin{figure*}[t]
    \centering
    \includegraphics[width=0.9\textwidth]{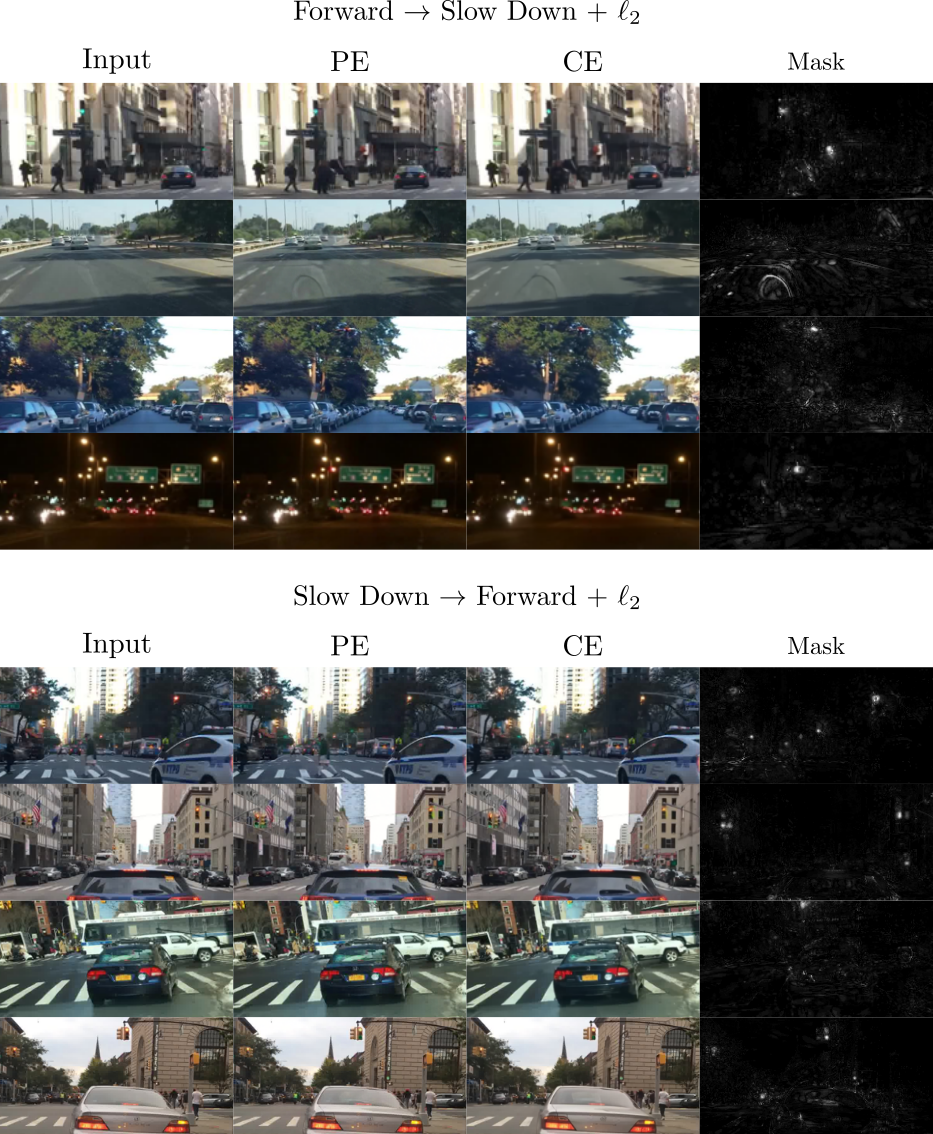}
    \caption{Additional BDD qualitative results. We show examples for the \emph{Forward / Slow Down} binary class for $\ell_2$ distance loss. We show a zoom of the changes in the image since the perturbations are sparse. We see that ACE adds traffic light colors in the buildings to change the prediction.}
    \label{soup:fig:celeba-smile}
\end{figure*}

\begin{figure*}[t]
    \centering
    \includegraphics[width=0.9\textwidth]{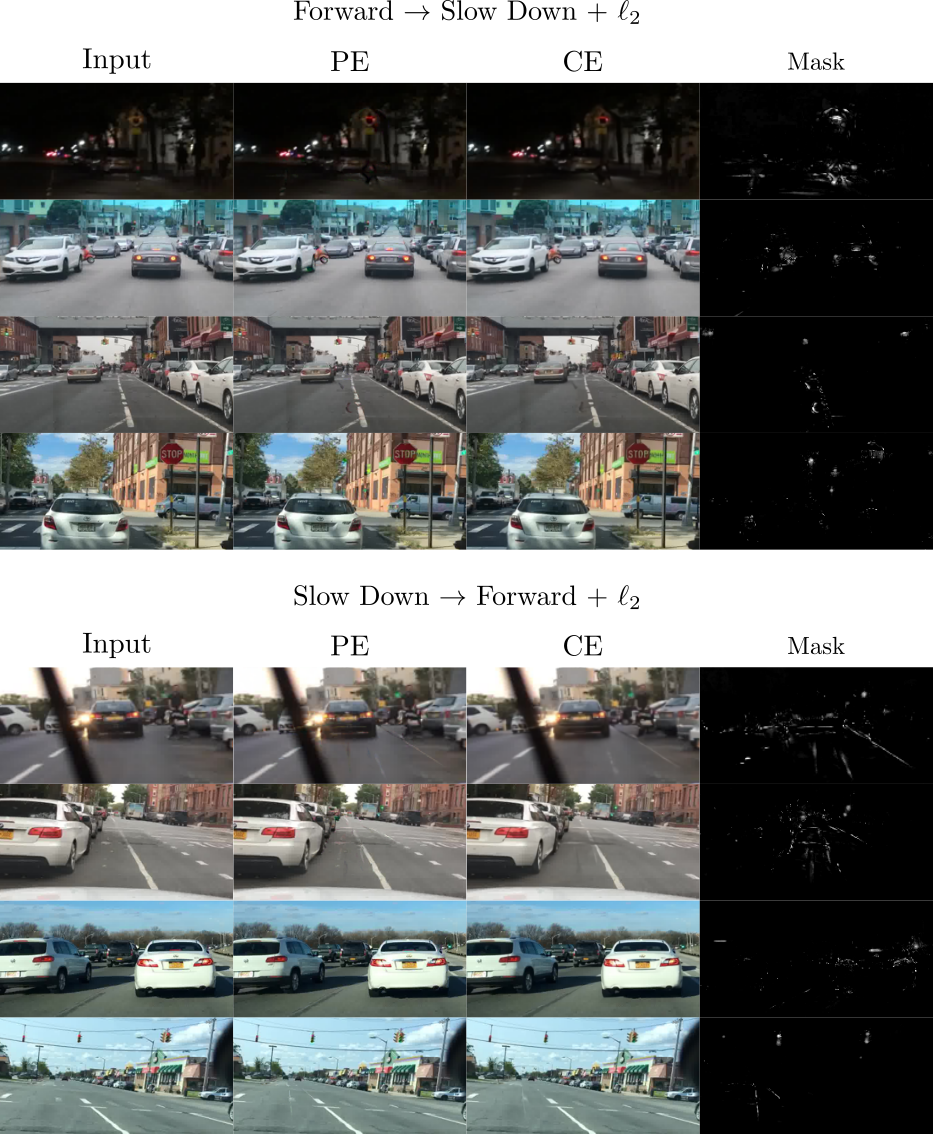}
    \caption{Additional BDD qualitative results. We show examples for the \emph{Forward / Slow Down} binary class for $\ell_1$ distance loss. We show a zoom of the changes in the image since the perturbations are sparse. We show a zoom of the changes in the image since the perturbations are sparse. We see that ACE adds traffic light colors in the buildings to change the prediction.}
    \label{soup:fig:celeba-smile}
\end{figure*}

\begin{figure*}[t]
    \centering
    \includegraphics[width=0.9\textwidth]{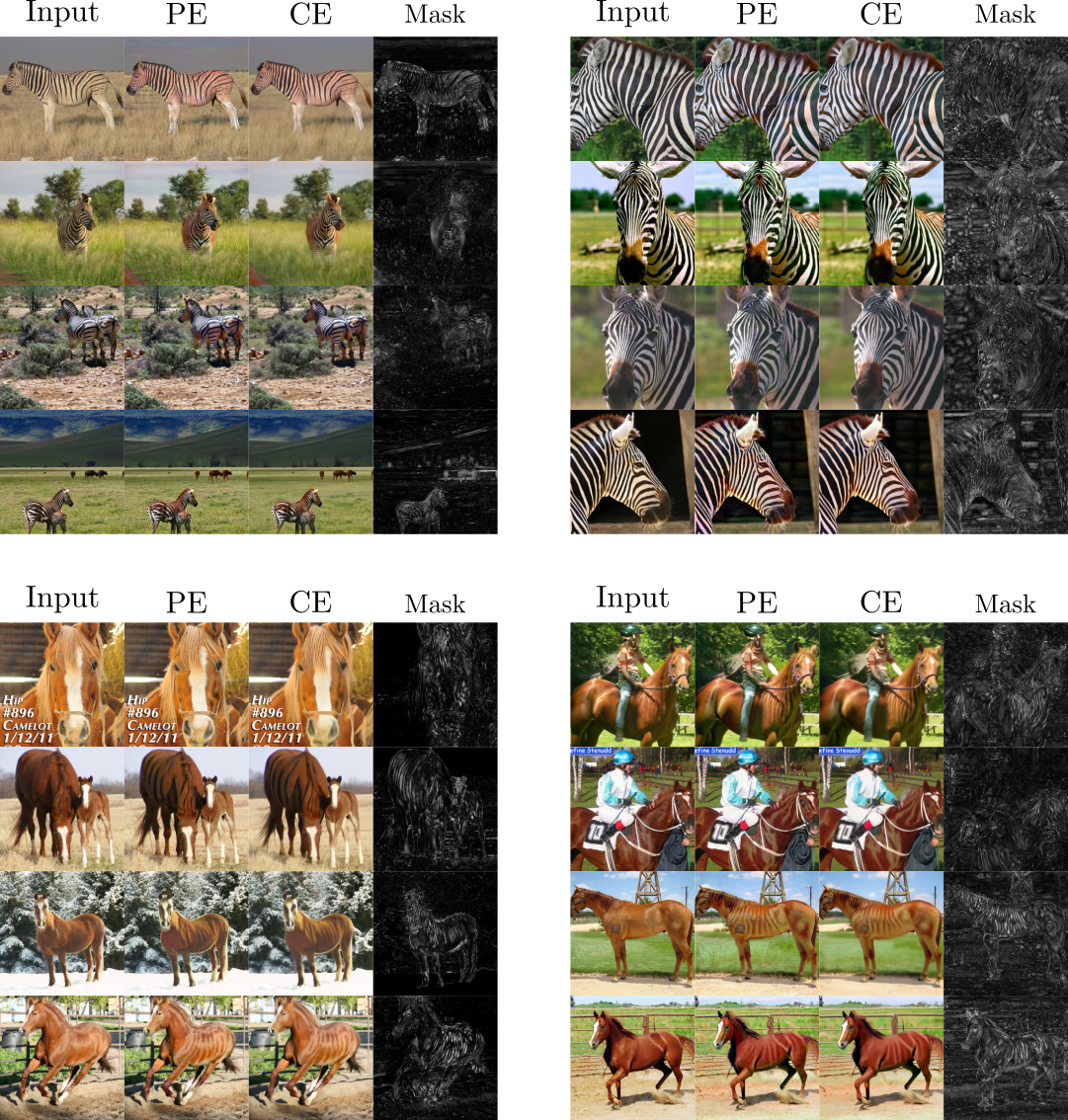}
    \caption{Additional ImageNet qualitative results. We show examples for the \emph{Zebra / Sorrel} categories class. The first column is the $\ell_1$ distance loss while the second one is $\ell_2$. The initial row is zebra to sorrel and the second one is the inverse. To change from zebras to sorrels, some examples show not only incorporating the brown color sorrel horses but also the context in the background (\eg adding a stable-like background). Vice-versa, to classify a horse as a zebra it is enough to add some strips.}
    \label{soup:fig:celeba-smile}
\end{figure*}

\begin{figure*}[t]
    \centering
    \includegraphics[width=0.9\textwidth]{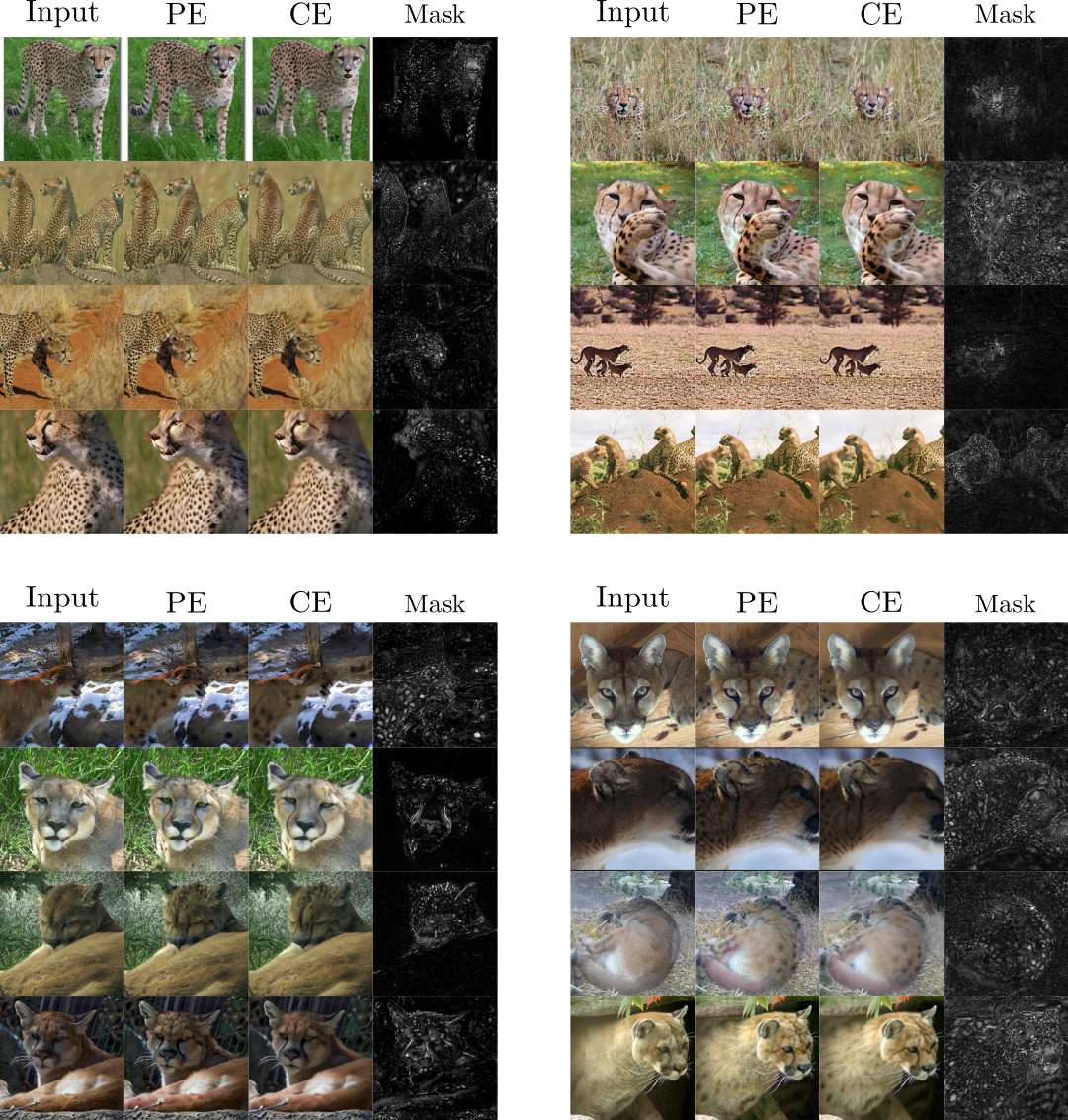}
    \caption{Additional ImageNet qualitative results. We show examples for the \emph{Cheetah / Cougar} categories class. The first column is the $\ell_1$ distance loss while the second one is $\ell_2$. The first row is cheetah to cougar and the second is the inverse. We mainly see that changing from cheetah to cougar is enough to target the face of the animal. Vice-versa, to classify a cougar as a cheetah, ACE adds spots and characteristic cheetah stripes on the face.}
    \label{soup:fig:celeba-smile}
\end{figure*}

\begin{figure*}[t]
    \centering
    \includegraphics[width=0.9\textwidth]{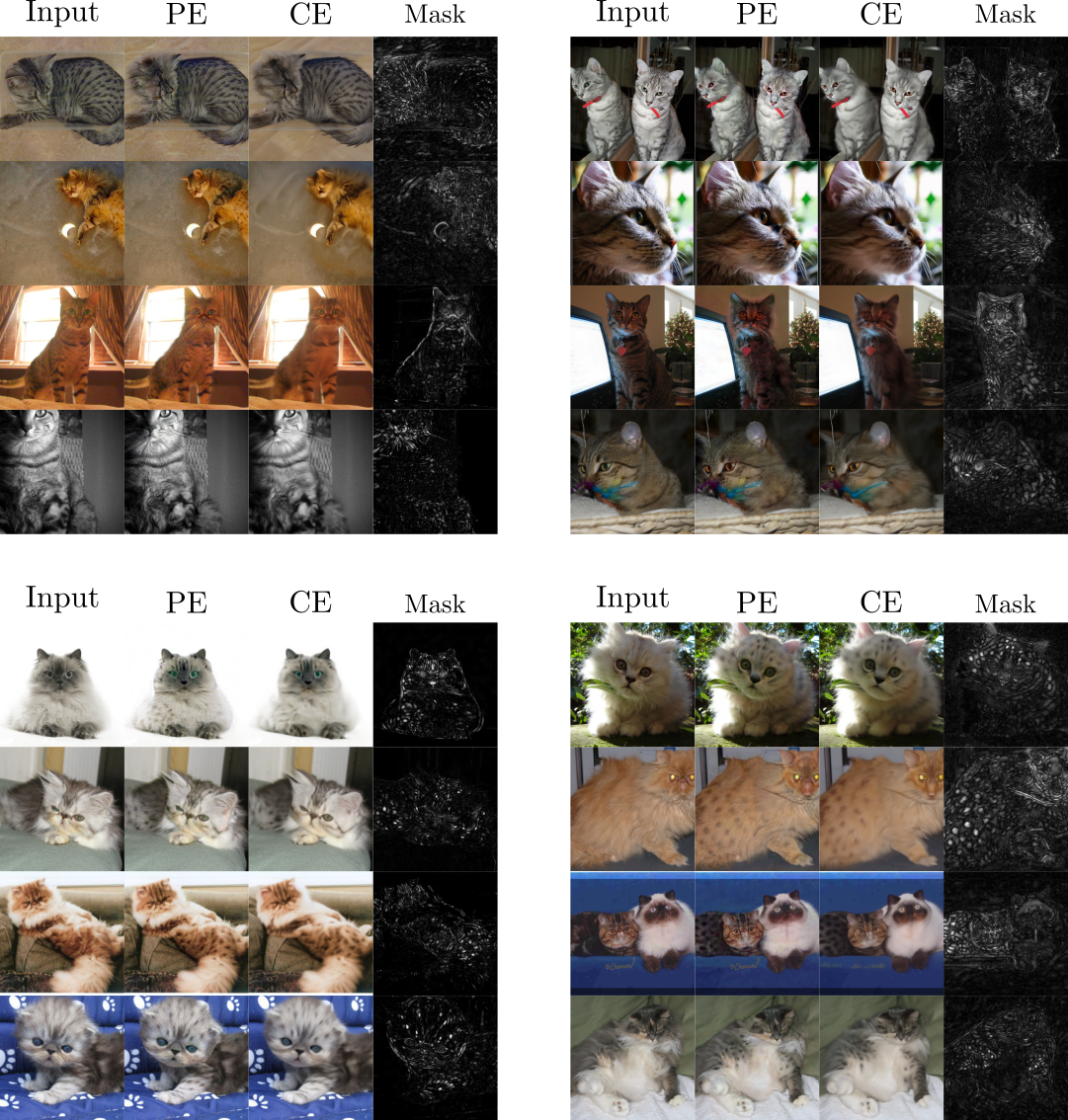}
    \caption{Additional ImageNet qualitative results. We show examples for the \emph{Egyptian / Persian cat} categories class. The first column is the $\ell_1$ distance loss while the second one is $\ell_2$. The row is Egyptian to Persian cat and the second is the inverse. To change from Egyptian to Persian, we mainly see that ACE adds the Persian cats' fluffy fur. Conversely, from Persian to Egyptian it adds spots.}
    \label{soup:fig:celeba-smile}
\end{figure*}

\end{document}